\documentclass[letterpaper, 10 pt, journal, twoside]{IEEEtran}

\usepackage[utf8]{inputenc}
\usepackage{amsmath,amssymb,stmaryrd,mathtools}
\usepackage{amsfonts}

\usepackage[noend]{algpseudocode}
\usepackage{acronym}
\usepackage{verbatim}
\usepackage{booktabs}
\usepackage{siunitx}
\usepackage{graphics}
\usepackage{graphicx,caption}
\usepackage{suffix}
\usepackage{xstring}
\usepackage{xparse}
\usepackage{expl3}
\usepackage{mathrsfs}
\usepackage{tabularx}
\usepackage{makecell}
\usepackage{array}
\usepackage{hyperref}
\usepackage{cleveref}
\usepackage[ruled,linesnumbered]{algorithm2e}
\usepackage{multirow}
\usepackage{diagbox}
\usepackage{rotating}
\usepackage{bbm}
\usepackage{dsfont}

\usepackage{subcaption}
\usepackage{wrapfig}

\usepackage{tikz}
\usetikzlibrary{arrows,backgrounds,calc}
\usepackage{relsize}
\usepackage{float}
\usepackage{kantlipsum} 
\usepackage{lipsum}
\usepackage{stfloats}
\usepackage{siunitx}

\usepackage[noadjust]{cite}
\usepackage{pifont}
\newcommand{\cmark}{\ding{51}}%
\newcommand{\circlemark}{\ding{109}}
\usepackage{todonotes}
\usepackage{soul}
\definecolor{smoothgreen}{rgb}{0.7,1,0.7}
\sethlcolor{smoothgreen}

\RequirePackage{luatex85}
\usepackage{pgfplots}
\pgfplotsset{compat=newest}
\pgfplotsset{every axis legend/.append style={%
		cells={anchor=west}}
}
\usepgfplotslibrary{polar}
\usetikzlibrary{arrows}
\tikzset{>=stealth'}

\definecolor{C1}{rgb}{0.0, 0.447, 0.741}
\definecolor{C1_light}{rgb}{0.0, 0.6032388663967612, 1.0}
\definecolor{C2}{rgb}{0.85, 0.325, 0.098}
\definecolor{C3}{rgb}{0.929, 0.694, 0.125}
\definecolor{C4}{rgb}{0.494, 0.184, 0.556}
\definecolor{C5}{rgb}{0.466, 0.674, 0.188}
\definecolor{C6}{rgb}{0.301, 0.745, 0.933}
\definecolor{C7}{rgb}{0.635, 0.078, 0.184}

\usepgfplotslibrary{groupplots}

\usetikzlibrary{shapes.geometric, arrows}

\tikzstyle{startstop} = [rectangle, rounded corners, minimum width=2cm, minimum height=1cm,text centered, draw=black, fill=none]
\tikzstyle{arrow} = [thick,->,>=stealth]

\newcommand{\etal}{\textit{et al}.}

\newcommand{\eg}{\textit{e}.\textit{g}.}
\newcommand{\modelName}{DRAGON\xspace}

\title{
DRAGON: A Dialogue-Based Robot for Assistive Navigation with Visual Language Grounding}

\author{Shuijing Liu, Aamir Hasan, Kaiwen Hong, Runxuan Wang, Peixin Chang, Zachary Mizrachi, \\Justin Lin, D. Livingston McPherson, Wendy A. Rogers, and Katherine Driggs-Campbell
\thanks{Manuscript received: July, 10, 2023; Revised November, 15, 2023; Accepted January, 21, 2024.}
\thanks{This paper was recommended for publication by
Editor Tetsuya Ogata upon evaluation of the Associate Editor and Reviewers’
comments. \textit{(Corresponding author:
Shuijing Liu.)}
}

\thanks{S. Liu, A. Hasan, K. Hong, R. Wang, P. Chang, Z. Mizrachi, J. Lin, D. McPherson and K. Driggs-Campbell are with the Department of  Electrical and Computer Engineering at the University of Illinois Urbana-Champaign. W. A. Rogers is with the Department of Applied Health Sciences at the University of Illinois Urbana-Champaign. emails: \texttt{\{sliu105, aamirh2, kaiwen2, runxuan6,pchang17, zdm3, jklin6, dlivm, wendyr, krdc\}@illinois.edu}}
\thanks{This work was supported in part by a Research Support Award from the University of Illinois Urbana-Champaign Campus Research Board, the National Science Foundation under Grant No. 2143435, and a grant from the National Institute on Disability, Independent Living, and Rehabilitation Research (NIDILRR) under Grant No. 90REGE0021.}%
\thanks{The user study was approved by IRB \#23565 at the authors' institution.}%

\thanks{Digital Object Identifier (DOI): 10.1109/LRA.2024.3362591.}
}

\begin{document}
\maketitle
\begin{abstract}
Persons with visual impairments (PwVI) have difficulties understanding and navigating spaces around them. 
Current wayfinding technologies either focus solely on navigation or provide limited communication about the environment.
Motivated by recent advances in visual-language grounding and semantic navigation, we propose \modelName, a guiding robot powered by a dialogue system and the ability to associate the environment with natural language.
By understanding the commands from the user, \modelName is able to guide the user to the desired landmarks on the map, describe the environment, and answer questions from visual observations. 
Through effective utilization of dialogue, the robot can ground the user's free-form language to the environment, and give the user semantic information through spoken language. 
We conduct a user study with blindfolded participants in an everyday indoor environment. 
Our results demonstrate that \modelName is able to communicate with the user smoothly, provide a good guiding experience, and connect users with their surrounding environment in an intuitive manner. 
Videos and code are available at {\color{cyan}{\url{https://sites.google.com/view/dragon-wayfinding/home}}}.

\end{abstract}

\begin{IEEEkeywords}
Human-Centered Robotics, Natural Dialog for HRI, AI-Enabled Robotics.
\end{IEEEkeywords}
\section{Introduction}
\label{sec:intro}
\IEEEPARstart{W}{ayfinding}, defined as helping people orient themselves in an environment and guiding them from place to place, is a longstanding challenge for persons with visual impairments (PwVI)~\cite{kulyukin2004rfid,bayles2022interdisciplinary}.
To improve the quality of PwVI's lives, we present a guiding robot that can connect language to the surrounding world to verbally interact with PwVI.

To pair wayfinding with communication, a line of previous works gives users signals such as navigation instructions~\cite{li2019ballbot,jin2021wearable} and basic environment information~\cite{wilson2007swan,guerreiro2019cabot}. 
As a step further, other wayfinding technologies recognize and convey the semantic meaning of the surrounding environment such as naming the landmarks~\cite{cheraghi2017guidebeacon,sato2017navcog3,kulyukin2006robot}.
However, these methods require special environmental setups, such as multiple RFID tags and bluetooth beacons.
To improve the aforementioned systems with recent advances in machine learning~\cite{chang2021learning,huang2022visual,ahn2022saycan}, we aim to remove dependence on these types of special infrastructure by integrating advances in visual-language grounding into conversational wayfinding.

More broadly, technologies in vision-language navigation and voice-controlled robots have made significant progress~\cite{chang2021learning,huang2022visual,ahn2022saycan}. 
These navigation agents are able to perform various tasks according to natural language commands such as ``bring me a cup'' with simple onboard sensors. 
This is usually achieved by encoding visual landmarks in a semantic map and associating language with these landmarks during navigation, which is referred to as visual-language grounding~\cite{huang2022visual,shah2022lmnav}.
However, these general-purpose frameworks assume that humans can provide step-by-step navigation instructions. 
These systems are not built for PwVI, who often need help perceiving the environment and planning paths. 
Thus, building a robot guide that can intuitively exchange semantic information with users remains an open challenge.

\begin{figure}[tb]
\centering
\vspace{5pt}
\includegraphics[scale=0.18]{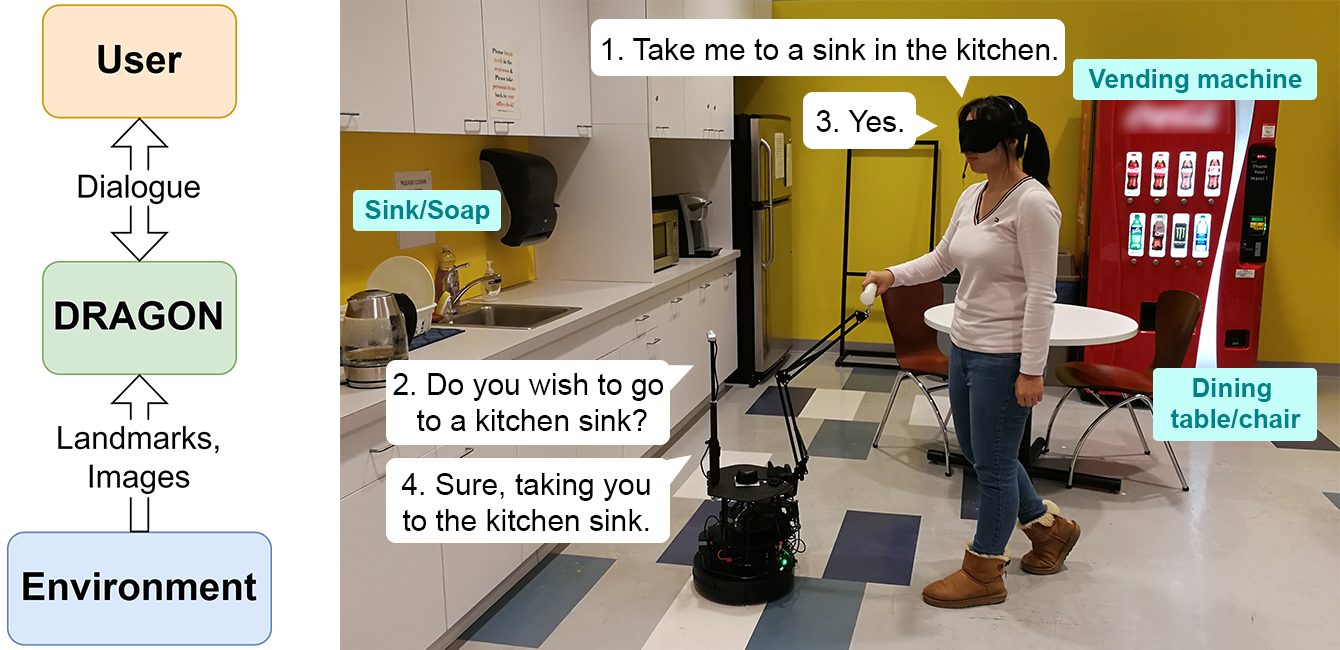}
\caption{\modelName identifies the intents of the user through dialogue, grounds language with the environment, and guides the user to their desired goal.}
\label{fig:overview}
\vspace{-12pt}
\end{figure}

In this paper, we propose \modelName, a \textbf{D}ialogue-based \textbf{R}obot for \textbf{A}ssistive navigation with visual-language \textbf{G}r\textbf{o}u\textbf{n}ding.
In Fig.~\ref{fig:overview}, 
since PwVIs have limited vision, \modelName uses speech to communicate with the user and a physical handle for fully autonomous navigation guidance. 
The dialogue and navigation can be executed simultaneously. 
When the user gives a speech command, Speech Recognition (SR) and Natural Language Understanding (NLU) modules first extract the user's intents and desired destinations. 
The user command does not have any templates or constraints on vocabulary or expressions.
Based on the outputs of NLU, one of the following grounding functionalities is triggered:
(1) finding users' desired destinations with a visual-language model~\cite{radford2021lclip} and guiding them to the destinations; 
(2) describing nearby objects; 
and (3) answering questions from users.
With (2) and (3), \modelName can help users gain awareness of their surroundings during navigation.

To find users' intended goals on a map, we propose a novel landmark recognition module based on CLIP~\cite{radford2021lclip}. 
After a straightforward mapping process, the landmark recognizer is able to select the landmark whose image best matches the user descriptions. 
Our landmark recognizer is able to associate flexible and open-vocabulary commands with few constraints on user expressions. 
If the description is ambiguous, our system will disambiguate user intents through additional dialogue.
Then, the corresponding goal location is passed to the path planners for navigation guidance. 
Combined with the robot's navigation module, the powerful and reliable landmark recognizer is essential to ensure the success and user experience of \modelName.



Our main contributions are as follows:
    (1) As an interactive navigation guide for PwVI, \modelName enables voice-based dialogue, which carries rich information and has grounding capabilities;
    (2) We propose a novel landmark mapping and recognition method that can associate free-form language commands with image landmarks. Our method can be easily plugged into the standard navigation module of mobile robots;
    (3) A user study with five blindfolded participants (N=5) demonstrates that \modelName is able to understand user intents through dialogue and guide them to desired destinations in an intuitive manner.  
To the best of our knowledge, our work is the first to show that visual-language grounding via dialogue benefits robotic assistive navigation.


\section{Related Works}
\label{sec:related}

\subsection{Wayfinding robots and technologies}
\label{sec:related_wayfinding}

\begin{table*}[hb]
\vspace{-5pt}
  \centering
    \caption{Benchmark for conversational wayfinding technologies. A \cmark\, means that the functionality is implemented. A \circlemark\, means partial implementation. A blank cell means the functionality is absent. (In \cite{kulyukin2006ergonomics}, the users have to enter a number sequence into a keypad to specify their destinations. \cite{saha2019closing} can only describe a fixed set of pre-mapped landmarks and can only answer two fixed questions.)}
    
    \label{tab:benchmark}
    \begin{tabular}{@{}lccccccccc@{}} 
    \toprule
    
     \multirow{2}{*}{\textbf{Method}} &
     \textbf{User-chosen}&
     \multicolumn{2}{c}{\textbf{Speech dialogue}} &
     \textbf{Environment} & 
     \multirow{2}{*}{\textbf{VQA}}& 
     \multirow{2}{*}{\textbf{Form}} & \textbf{Environmental}\\
     \cmidrule{3-4}
     & \textbf{semantic goals}& Input & Output & \textbf{description}& & & \textbf{Instrumentation}\\
   \midrule
  
   GuideBeacon~\cite{cheraghi2017guidebeacon} &\cmark&\cmark&\cmark&&&Phone application & Bluetooth beacons \\
   NavCog3~\cite{sato2017navcog3}&\cmark&\cmark&\cmark&\circlemark&\circlemark& Phone application& Bluetooth beacons\\
   LandmarkAI~\cite{saha2019closing} &\cmark&\cmark&\cmark &\cmark&\circlemark& Phone application& GPS\\ 
    SeeWay~\cite{yang2022seeway} &\cmark & \cmark &\cmark&&&Phone application& WiFi \\
     Robotic Shopping cart~\cite{kulyukin2006ergonomics}&\circlemark&&\cmark&&& Robot & RFID tags \\
    CaBot~\cite{guerreiro2019cabot} &&&\cmark&\cmark&&Robot& Remote joystick\\
    Ballbot~\cite{li2019ballbot}&\cmark&\cmark&\cmark&&&Robot & WiFi + Remote computer \\
    Ours &\cmark&\cmark&\cmark&\cmark&\cmark& Robot & WiFi + Remote computer\\
      \bottomrule
    \end{tabular}
\end{table*}

\textbf{Navigation guidance:} To guide PwVI from point A to point B following a planned path, unactuated devices, such as smartphones and wearables, rely on haptic or audio feedback to give instructions such as going straight and turning right~\cite{jin2021wearable,yang2022seeway,cheraghi2017guidebeacon,sato2017navcog3}.
However, delays and misunderstandings might lead to inevitable deviations, which take time and effort to recover from~\cite{wilson2007swan,cheraghi2017guidebeacon}.
On the other hand, robots provide a physical holding point, which offers kinesthetic feedback to minimize deviations and reduce the mental load of users~\cite{kulyukin2006ergonomics,nanavati2019coupled_dyads,zhang2023follower}. 
Such physical guidance can be combined with aforementioned verbal or haptic navigation instructions to further improve performance at the cost of a more expensive system~\cite{guerreiro2019cabot,li2019ballbot}. 
To ensure both efficiency and low cost, we mount a handle on our robot to give intuitive real-time steering feedback in navigation. 


\textbf{Semantic communication:} 
A large part of blind navigation technologies ignores exchanging environmental information with users~\cite{li2019ballbot,nanavati2019coupled_dyads,xiao2021guidedog}. 
To deal with this issue,
CaBot applies object recognition to describe the user's neighborhood, yet the user cannot hold conversations with the robot or choose their destinations~\cite{guerreiro2019cabot}. 
To enable users to choose a semantic goal (\eg\: a restroom), some works mark points of interest using bluetooth beacons~\cite{cheraghi2017guidebeacon,sato2017navcog3} or RFID tags~\cite{kulyukin2006robot,kulyukin2006ergonomics}, which requires heavy instrumentation. 
As an alternative, extracting semantic information from ego-centric camera images is much cheaper and easier. 
For example, SeeWay uses skybox images to represent landmarks~\cite{yang2022seeway}. 
Similarly, Landmark AI offers semantic-related functionalities including describing the environment, reading road signs, and recognizing landmarks using a phone camera~\cite{saha2019closing}. 
However, these phone applications are not robots and thus cannot physically guide users or provide a stable mounting point for cameras. In contrast, Table~\ref{tab:benchmark} shows that \modelName brings conversational wayfinding to the next level: A robot can simultaneously offer physical guidance and enable users to trigger various functionalities through dialogue. 



\subsection{Command following navigation}
\label{sec:related_command_follow}
Tremendous efforts have been made in understanding and grounding human language instructions for various robotic tasks~\cite{chang2021learning,ahn2022saycan,huang2022visual}. 
In command following navigation, a modular pipeline usually consists of three  modules: 
(1) an NLU system to map instructions to speaker intent; 
(2) a grounding module to associate the intent with physical entities; and 
(3) a SLAM and a planner to generate feasible trajectories~\cite{liu2019review,min2021film,shah2022lmnav}. 
Other works attempt to learn end-to-end policies from simulated environments or datasets~\cite{anderson2018vision, chen2020soundspaces,chang2020robot,gadre2022cow}. However, due to sim-to-real gaps in perception, language, and planning, deploying these policies to the real world remains an open challenge for applications in the low data regime such as wayfinding~\cite{Zhu2020The}. 
Therefore, we adopt the modular pipeline to ensure performance in the real world. 

\subsection{Semantic landmark recognition}
Understanding the semantic meanings of a scene is a vital step towards interactive navigation~\cite{huang2022visual,shah2022lmnav}. 
Some works reconstruct volumetric maps for the environment, where each grid is associated with a semantic label~\cite{mccormac2017semanticfusion,min2021film,huang2022visual}. Other works build more abstract scene graphs~\cite{hughes2022hydra,wu2021scenegraphfusion}.
However, implementing these methods on a real robot is expensive, as they require accurately calibrated depth cameras and high-performing instance segmentation models. 

Another line of work collects images as landmarks to create topological graphs~\cite{savinov2018semiparametric,meng2020scaling,chaplot2020neural}. In navigation, the goal location is retrieved by computing the similarity between a goal image and all stored landmarks. 
However, the above works only consider image goals, which are less natural than language in human-centered applications. 
Inspired by Shah \etal~\cite{shah2022lmnav} and Huang \etal~\cite{huang2022visual}, we use CLIP~\cite{radford2021lclip} to associate image landmarks with users' language commands. 
Compared with previous works that use closed vocabulary object detectors, which are limited to a predefined set of semantic classes \cite{mccormac2017semanticfusion,wu2021scenegraphfusion,hughes2022hydra}, our method can handle more flexible and open-vocabulary commands.
We use CLIP to select landmarks and keep traditional cost maps for planning, enabling easy integration of our method into the navigation stack of mobile robots.

\section{System Overview}
\label{sec:system-overview}

\begin{figure*}[!ht]
    \centering
    \includegraphics[scale=0.45]{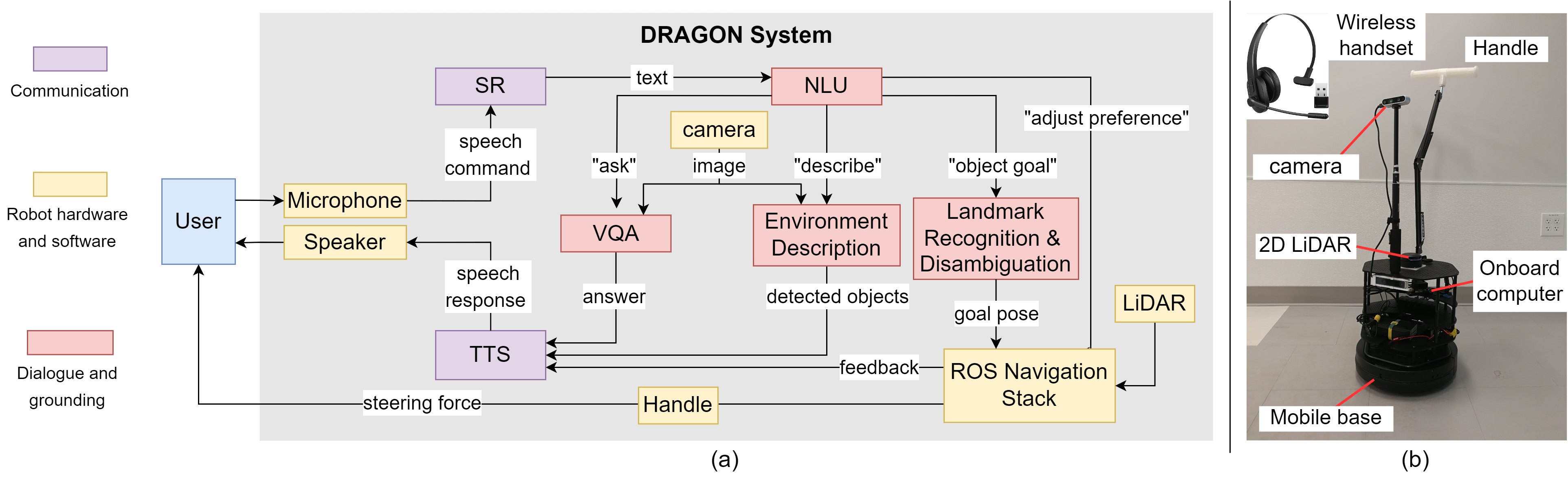}
    \vspace{-3pt}
    \caption{\textbf{An overview of the system and platform of \modelName}. (a) Submodules, message passing, and user interface. (b) The robot platform.}
    \label{fig:system}
    \vspace{-13pt}
\end{figure*}
In this section, we describe the setup and configuration of our robot guide with special considerations for PwVI users. 
Fig.~\ref{fig:system}(a) shows an overview of our proposed system with three main components: (1) The TurtleBot platform (yellow); (2) Audio communication interface (purple); (3) Dialogue and grounding modules (red). The modules communicate with each other through ROS.
We expand part (1) and (2) in this section and part (3) in Sec.~\ref{sec:semantic-dialogue}.
\subsection{Robot platform}
\textbf{Overview:}
\label{sec:overview-turtlebot}
We use the Turtlebot2i as our robot platform.
As shown in Fig.~\ref{fig:system}(b), the robot is fitted with the following sensors and equipment:
(1) An RP-Lidar A3 laser range finder is mounted on the top of the robot structure for SLAM;
(2) An Intel RealSense D435i camera is mounted on the top of a monopod facing forward for scene description and question answering;
(3) A wireless headset is used to communicate with the user. The headset is lightweight and maximally protects the users' privacy, while the absence of wires avoids tripping hazards; 
(4) A T-shaped handle is attached to the top rear side of the robot as a holding point for the user's arm.
The handle allows users to choose their preferred holding configurations such as one hand or two hands.
The robot is connected to a desktop computer which provides more computation resources through WiFi. 

\textbf{Planning and Navigation:}
\label{sec:overview-planning}
The robot operations are managed by the ROS 
\texttt{move\_base} navigation stack, which is a standard package to autonomously navigate a mobile robot to a given goal pose.
Before navigation, we create a 2D occupancy map of the environment using laser-based SLAM and mark the semantic landmarks at the same time (see Fig.~\ref{fig:environment} and Sec.~\ref{sec:clip} for details).
At the beginning of each trial, the goal pose is obtained from the dialogue with the user (further specified in Sec.~\ref{sec:semantic-dialogue}). 
During navigation, adaptive Monte Carlo localization is used to localize the robot on the map.
We use the dynamic window approach (DWA)~\cite{fox1997dynamic} and $\text{A}^*$ as local and global planners, respectively. 
The minimum translational velocity is restricted to be non-negative to prevent the robot from moving backward and colliding with the user. 
The maximum velocity of the robot can be adjusted by the user (see Sec.~\ref{sec:nav_speed}).

\subsection{Audio Communication Interface}
\label{sec:overview-communication}
Speech is a natural choice for human-robot communication, particularly in cases where the human has limited vision~\cite{azenkot2016enabling,bhat2022confused,bayles2022interdisciplinary}. 
To this end, as shown in purple in Fig.~\ref{fig:system}(a), we develop an audio communication interface that consists of: 
(1) Input: When audio is captured by the \texttt{audio\_capture} package, the OpenAI Whisper speech recognition model~\cite{whisper} transcribes speech commands to text, which are passed to the NLU module. 
The SR module continuously transcribes the audio from the microphone and publishes the text transcriptions to a ROS topic in real time;
(2) Output: We use the Google text-to-speech (TTS) service to convert the text output from the visual-language modules and navigation module to speech, which is then narrated to the user via the headset. 
The TTS is another ROS topic that converts and plays the synthesized sound constantly.



\section{Dialogue And Grounding}
\label{sec:semantic-dialogue}

The goal of \modelName is to connect the user with the environment through conversation. 
In this section, we describe how our dialogue system understands user language~(Sec.~\ref{sec:nlu}), maps and localizes semantic landmarks~(Sec.~\ref{sec:clip}), provides information about the environment~(Sec.~\ref{sec:img_caption_vqa}), and adjusts the navigation preference of the user~(Sec.~\ref{sec:nav_speed}). 
Our grounding system is visualized in the red parts of Fig.~\ref{fig:system}(a). The inputs to the subsystem are the transcribed texts from SR and the outputs are synthetic speech from TTS.

\subsection{Natural language understanding (NLU)}
\label{sec:nlu}
The NLU takes a transcribed sentence as input and outputs user intents and entities of interest. The intent recognizer is a multi-label classifier with all classes shown in Table~\ref{tab:intents}. The intents are designed based on the needs of our tasks.
The entities are locations, objects, and object attributes which include the material and functionalities of an object.  
We use Dual Intent and Entity Transformer for intent classification and entity recognition~\cite{bunk2020diet}. We train the model using a custom dataset with $1092$ sentences collected by ourselves. 
For each intent, we collect various expressions including misspelled and phonetically similar phrases, which makes our NLU robust to the nuances of human language and the errors caused by the SR. 
For example, ``a think'' and ``a sink'' both refer to the kitchen sink. 
We also collected expressions for multi-intents and unknown intents so that the NLU can fulfill a request containing multiple intents and ignore noise input. 
For instance, ``Hello robot, can you take me to a sofa?'' will both activate the robot and set an object goal. 
Once the intent and entities are extracted, the corresponding downstream module is activated. 
The NLU may pass additional input arguments to modules such as extracted entities or the whole sentence.
Different downsteam models are triggers based on extracted intents and entities.

During navigation, the landmark recognition is triggered if the user intent is \textit{Object goal} or \textit{Location goal} and the NLU extracts a goal object from the input sentence. The extracted information of the goal is kept in memory throughout the conversation. 
If the user mentions additional information about the landmark, we use simple prompt engineering to make the description more specific. 
For example, locations and attributes of objects, such as  ``a chair in the kitchen,'' can be added to the memorized description.
In addition, the robot uses clarification dialogue to disambiguate the desired landmark if the input description does not contain any object. 
If the user only provided the location or attributes without mentioning the object name (\eg~``Take me to the kitchen''), our system provides hints to encourage the user to provide more specific descriptions (\eg~``What object are you looking for in the kitchen?'').
If there are multiple similar objects in different landmarks, our system disambiguates the user's preferred landmark (\eg~``What kind of chair are you looking for? A dining chair, an office chair, or a sofa?'').
After choosing a unique landmark, our system confirms the memorized goal description with the user (\eg~``Do you wish to go to a dining chair?'').
No further action is taken until the user affirms the goal. The memorized goal information is cleared after the confirmation to prepare for the next goal.

With the disambiguation and confirmation dialogue, the NLU is able to precisely capture the user's desired destination with minimal constraints on the user's phrasing, which is crucial for the whole navigation experience. 
Using better language models for the NLU is left for future work.

\begin{table}[!tb]
  \begin{center}
    \caption{All user intents and their descriptions.}
    \label{tab:intents}
    \begin{tabular}{@{}ll@{}} 
    \toprule
    
     \multirow{1}{*}{\textbf{Intents}} &
     \multirow{1}{*}{\textbf{Descriptions}} \\
     \midrule

      Greet & Wake up the robot and begin an interaction. \\

      Object goal & Go to a specific object landmark. \\&May contain entities including objects and attributes.\\

      Location goal & Go to a rough goal location (kitchen, lounge, etc) \\& without mentioning a specific object.\\
      & May contain location entities.\\

      Affirm & Confirm the goal. \\

      Deny & Deny the goal. \\

      Describe & Ask for a description of the surrounding environment. \\

      Ask & Ask a question about the surrounding environment. \\

      Pause & Pause the current navigation.\\

      Resume & Resume the current navigation.\\

      Accelerate & Increment velocities, up to a limit.\\

      Decelerate & Decrement velocities, down to a limit.\\

      Unknown & The text does not belong to any intents above (i.e. \\ & be noise, chitchat, etc) and is ignored by the robot.\\
      \bottomrule
    \end{tabular}
  \end{center}
  \vspace{-20pt}
\end{table}



\subsection{Landmark mapping and recognition}
\label{sec:clip}
To guide the user to their object goals, we first record the images and locations of landmarks during SLAM. 
Then, we use a fine-tuned CLIP model to match the user's description with goal images, whose corresponding location and orientation are sent to the navigation stack for navigation guidance.
The CLIP model version is ViT-B/32~\cite{radford2021lclip}.

The landmark mapping process is performed simultaneously with SLAM. 
During SLAM, when the robot is at a landmark that might be a point of interest, we simply save the current robot pose in the map frame and an RGB image of the landmark to the disk with a single key press. 
No labels or text descriptions are needed at this stage.
The resulting landmark map is shown in Fig.~\ref{fig:environment}.

During navigation, this module is activated when the intent is \textit{Object goal} or \textit{Location goal}. 
After the goal is confirmed by the NLU, the CLIP model selects the landmark whose image has the highest similarity score with the descriptions of landmarks. 
To obtain the image-text similarity score, a text encoder and an image encoder first convert the input text and all images to vector embeddings.
Then, the text and image similarity score is computed by the cosine similarity between the pairwise text and image embeddings. 
The image with the highest similarity score is selected as the goal. 
Finally, the corresponding location of the chosen landmark on the map is sent to an action client, which sets the goal for the robot.

The zero-shot performance of pre-trained CLIP models is not satisfactory in our environment due to distribution shifts. 
As shown in Fig.~\ref{fig:environment}, the objects in the images are frequently cropped due to the low mounting point of our camera and the close distance between the camera and the objects. 
In addition, the descriptions of landmarks from a PwVI might be vaguer than those in public datasets (\eg~``a chair'' v.s. ``a blue chair in front of a white wall'').
To this end, we fine-tune the CLIP model with a custom dataset containing $544$ image and text description pairs with a $8\times 10^{-6}$ learning rate for 35 epochs.
The images are taken by the robot camera in our environment and the text is provided by the authors.
By using an open-vocabulary model to recognize landmarks, \modelName can handle free-form language and is not limited to a fixed set of object classes. Thus, the user expressions are less restricted, making the grounding module easier for non-experts to use.

\subsection{Environment understanding modules}
\label{sec:img_caption_vqa}
To help the user gain awareness of their surroundings, 
we use an object detector~\cite{zhou2022detecting} to describe the objects (activated if the intent is \textit{Describe}) and a VQA model~\cite{kim2021vilt} to answer the user's questions (activated if the intent is \textit{Ask}). 
Both models take the current camera image as input. 

The output of the object detector consists of a list of detected instances, their object classes, confidence scores, and bounding boxes. 
To avoid narrating a long list and to keep the description concise, we post-process the output as follows. 
We first apply non-maximum suppression and filter out the detected instances with low confidence scores. 
Then, for the remaining instances, we keep the top three classes with the largest average bounding boxes, and list the object class names together with the numbers of objects (\eg~``2 chairs, 1 person, and 1 table''). 

The VQA model takes the current camera image and the user's question from the SR and outputs a short answer to the question. 
Since healthy people and PwVI would ask different questions to the same images~\cite{antol2015vqa}, we collect a dataset of $10252$ (image, question, answer) triplets to fine-tune the VQA model for $20$ epochs. Again, images are taken by the robot camera
in our environment and the text is provided by the authors.
To handle free-form user expressions, the dataset contains cases where multiple questions have the same meaning but different phrasing (\eg~``Is any person in front of me?'' and ``Anyone here?'').  

Finally, the outputs of the object detector and VQA are narrated to the user in real time. 
Since both models can only take an RGB image, our system cannot provide depth-based information or detect anything out of the camera view. 

\subsection{Navigation preference customization}
\label{sec:nav_speed}
To accommodate the different walking paces of users and to avoid tiring the user during navigation, the robot can change its speed (activated if the intent is \textit{Accelerate} or \textit{Decelerate}), take a pause (\textit{Pause}), and resume (\textit{Resume}). 
To pause the robot, our system stores and cancels the current goal from the action client in the navigation stack. 
To resume, the stored goal is sent to the action client again. 
To update the speed, we change the maximum translational and rotational velocities of the DWA local planner in real-time.

\section{Experiments}
\label{sec:experiments}
\subsection{Baseline}
We compare the CLIP-based landmark recognizer with a closed-vocabulary object detector as the baseline~\cite{zhou2022detecting}~\footnote{Open-vocabulary object detectors exist~\cite{zhou2022detecting}. We choose a closed-vocabulary detector to represent a closed-vocabulary grounding model.}. The vocabulary size, or the number of classes, of the detector is more than $1200$ and it is fine-tuned with the same amount of data as CLIP.
In the baseline, the landmark images are passed into the object detector, which outputs the class names of detected objects. 
During navigation, the baseline chooses the landmark with the highest number of objects mentioned by the user. 
Since the vocabulary of object detectors is fixed,
the baseline is unable to incorporate an object's attributes or locations obtained from disambiguation.  
All other modules are the same for our system and the baseline.

\subsection{User Study}
\begin{figure}[tb!]
    \centering
    \includegraphics[width=\columnwidth]{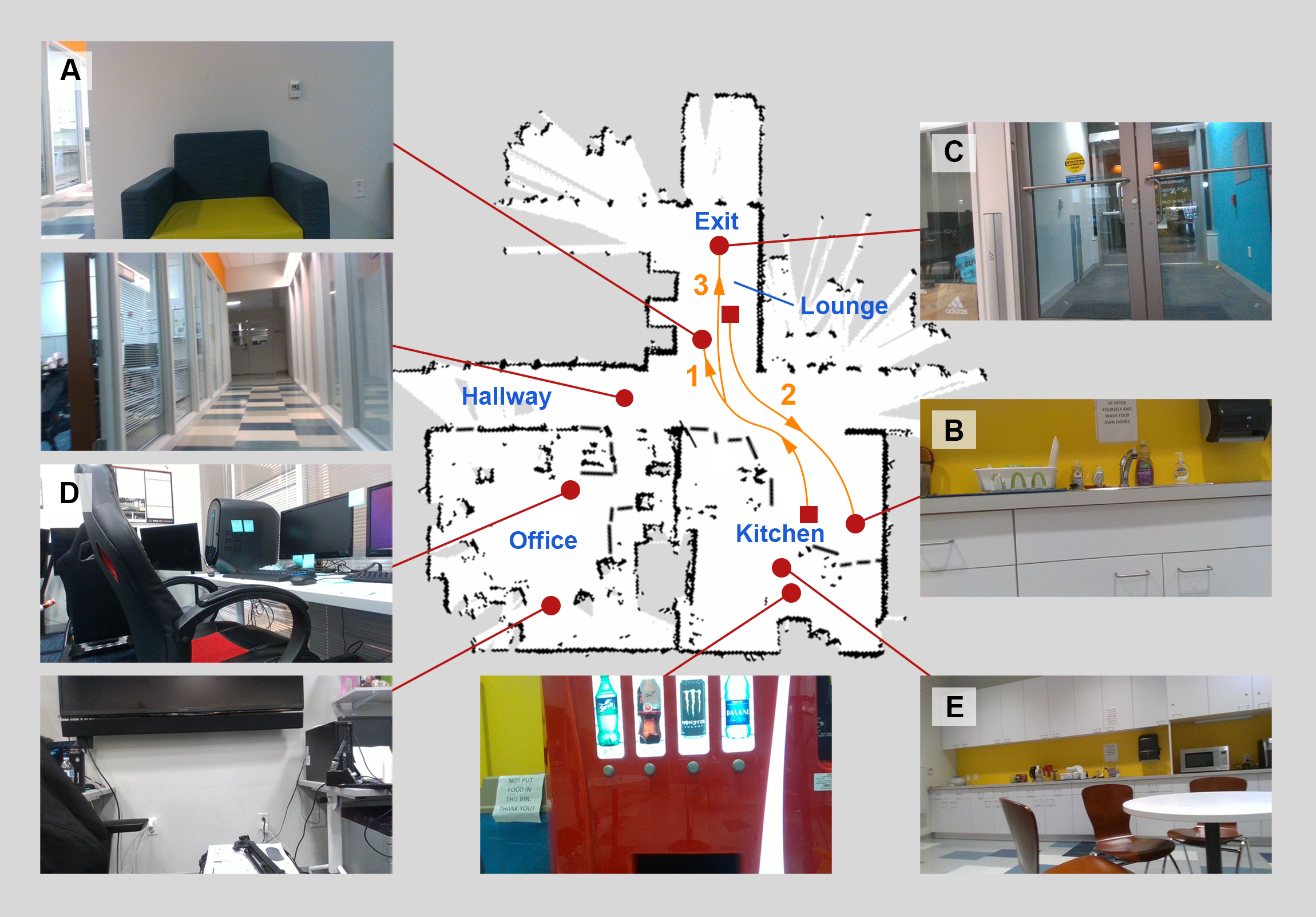}
    \caption{\textbf{The map of our environment with semantic landmarks.} The images are landmarks with locations marked by red dots. The orange lines are the three routes in the user study. The red squares are the starting locations of routes.} 
    \label{fig:environment}
    \vspace{-4pt}
\end{figure}

\textbf{Environment:} 
All experiments were conducted in an everyday indoor environment in a university building.
Three routes were created with furniture obstacles. 
Fig.~\ref{fig:environment} provides a layout of the environment, all landmarks, and three routes highlighted with orange curves.
The routes were designed to have varying levels of difficulties for the system to correctly interpret the destination. 
Specifically, landmark A of Route 1 contains simple objects, landmark B of Route 2 contains more complicated objects, and landmark C of Route 3 contains a transparent door that is hard for object recognition.

\textbf{Participants:}
The user study was conducted with N=5 participants (mean age=26; 3 males; 2 females; all participants were university students).
All participants have full (corrected) vision and are asked to wear a blindfold to simulate a visual impairment.
While our true target population are PwVI, the purpose of this pilot study is to validate the capabilities of \modelName.
A user study with PwVI is left for future work.

\begin{table}[t!]
  \begin{center}
    \caption{Example expressions and their corresponding landmarks from CLIP v.s. the detector. The landmark labels are from Fig.~\ref{fig:environment}. Underlined expressions are collected from the user study.}
    \label{tab:grounding}
    \begin{tabular}{c l l} 
    \toprule
    \textbf{Landmark} &  \multicolumn{1}{c}{\textbf{CLIP}}  & \textbf{Detector}  \\
    \midrule
    \multirow{3}{*}{A} & \underline{sofa}, \underline{couch}, \underline{coach},  &  \underline{sofa}, \\
      
     & fabric chair, relaxing chair &   thermostat \\
     & thermostat, climate control& \\
     
    \midrule
     \multirow{4}{*}{B} & \underline{sink}, \underline{think},\underline{sync}, \underline{faucet} & \underline{faucet} \\
     & \underline{soap}, hand wash, water pipe &  bottle \\
     & paper towel dispenser, bowls &  dispenser\\
     & kitchen countertop, drying rack & bowl\\
     \midrule
      \multirow{2}{*}{C} & \underline{door}, \underline{exit},  \underline{entrance}, gate & poster \\
     & glass door, automatic door &  \\
     
      \bottomrule
    \end{tabular}
  \end{center}
  \vspace{-17pt}
\end{table}

\textbf{Procedure:}
Participants were first familiarized with the goals of the study and requested to fill a demographic and robot technology survey.
Then, participants were provided with a test run to get familiar with the system and its intricate navigation feedback mechanism.
To begin the trial, the users were asked to command the robot to take them to a predetermined goal destination. 
Participants were not constrained in either the vocabulary or the sentence structure of their speech commands.
The users were also informed that they could interact with the robot (\eg~ask for a description of their surroundings) at any point of the navigation.
After each route, we used a short questionnaire
to measure the participant's perception of the system.
A strictly structured post-survey interview was conducted after participants finished all three routes to collect their feedback with the system.
The same procedure was performed for CLIP and the detector, resulting in a total of 15 trials per method (3 routes and 5 users). 
The order of which method was tested first was randomized for each participant to minimize the bias introduced due to the order of testing.
All materials included in the user study, including a full walkthrough of the whole study for a participant and all questionnaires can be found here: \url{https://drive.google.com/file/d/15KNR6C82mUrKSPMFRCnAJZ1C2NGX7dXJ/view}. 




\subsection{Metrics}

\textbf{Objective Metrics:}
We measure the accuracy of all interactions during the user study, including $312$ NLU, $30$ landmark recognition (LR) and navigation trials, $15$ environment description (EnvDes), $74$ VQA, and $15$ navigation preference adjustment (NavAdj). 
The NLU is correct when the extracted intent and entities (if any) are both correct. 
We also measure the accuracy of the NLU by taking the correctness of SR into account to analyze the effect of wrong SR. 
The effect of wrong NLU outputs is ignored when evaluating its downstream modules.
An LR is considered correct if the robot chooses the correct landmark. A navigation trial is successful if the robot guides the user to the correct landmark without any delays or collisions along the route.
An EnvDes is considered fully correct if all named objects exist in the camera image and the number of all objects is correct. It is considered partially correct if all named objects exist but the number of some objects is wrong.
The correctness of answers from VQA is based on the camera images, not on the information out of the camera view.
A NavAdj is successful if the change in robot speed is consistent with the user commands.

\textbf{Subjective metrics:}
For both methods, we compare the scores for categories from the short questionnaire in Table~\ref{tab:routes}.
The difference in scores for each participant was aggregated and analyzed to discount individual biases.
We evaluate user preferences for the other modules through a simple Likert scale analysis on the responses from the post-survey interview.
Additionally, participants' feedback is summarized for qualitative analysis.

\begin{table*}[b]
\vspace{-7pt}
    \begin{minipage}{0.4\linewidth}
    \centering
    \caption{Success rates (\%) of LR and navigation (including overall success rate, and success rate if LR is correct), and the average number of dialogue rounds for a successful LR.}
    \label{tab:dialogue_metrics1}
    \begin{tabular}{l c c c c} 
    \toprule
    \multirow{2}{*}{\textbf{Method}}&  \multicolumn{2}{c}{\textbf{LR}} & \multicolumn{2}{c}{\textbf{Navigation}}\\
    \cmidrule(lr){2-3}\cmidrule(lr){4-5}
      & Overall & \# rounds & Overall & Correct LR \\
    \midrule
    Ours & 100 & 2.4 &100&100\\
    Baseline &  46.67 & 3.71 & 46.67 & 100 \\
      \bottomrule
    \end{tabular}
    \end{minipage}\hfill
    \begin{minipage}{0.59\linewidth}
    \centering
    \caption{Accuracies (\%) of the SR, NLU (including overall accuracy, accuracy if SR is correct and if SR is wrong), EnvDes with fully correct and partially correct number of objects, VQA, and navigation adjustment modules.}
    \label{tab:dialogue_metrics2}
    \begin{tabular}{c c c c c c c c} 
    \toprule
    \multirow{2}{*}{\textbf{SR}} & \multicolumn{3}{c}{\textbf{NLU}} & \multicolumn{2}{c}{\textbf{EnvDes}} & \multirow{2}{*}{\textbf{VQA}} & \multirow{2}{*}{\textbf{NavAdj}} \\
    \cmidrule(lr){2-4}\cmidrule(lr){5-6}
      & Overall & Correct SR & Wrong SR  & Full & Partial & & \\
    \midrule
    \multirow{2}{*}{70.19} & 
    \multirow{2}{*}{85.26} & 
    \multirow{2}{*}{93.61} & 
    \multirow{2}{*}{65.59} & 
    \multirow{2}{*}{45.45} & 
    \multirow{2}{*}{75.76} & 
    \multirow{2}{*}{82.43} & 
    \multirow{2}{*}{100} \\
    \\


    \bottomrule
    \end{tabular}
    \end{minipage}
\end{table*}

\begin{figure*}[t!]
    \centering
    \includegraphics[width=\linewidth]{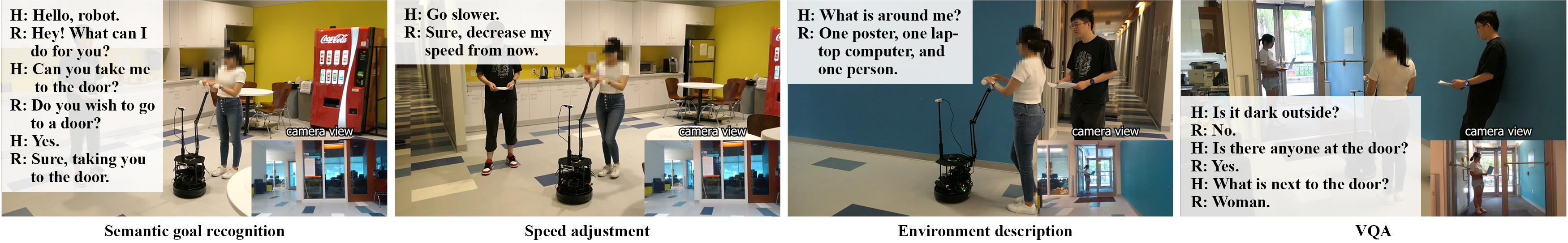}
    \caption{\textbf{An example navigation trial with human-robot dialogue in the user study.} In the dialogue boxes, ``H'' denotes the human and ``R'' denotes the robot. The camera view is shown in the lower right corner. }
    \label{fig:real_before_after}
    \vspace{-10pt}
\end{figure*}

\section{Results}
\label{sec:results}
In this section, we discuss the results of our user study.
Example navigation trials, as well as demonstrations of each module during the user study, are in \href{https://www.youtube.com/watch?v=1fojc44GTtI}{this video} and Fig.~\ref{fig:real_before_after}.

\subsection{Quantitative Evaluation}

\textbf{Landmark recognition and navigation:}
CLIP and the baseline only differ in LR and its resulting navigation.
As seen in Table~\ref{tab:dialogue_metrics1}, the success rate of navigation is 100\% if LR succeeds, 
because ROS navigation stack can navigate the robot to any desired goal pose robustly in our environment.
This dependency indicates that the performance of LR is the key factor for navigation in the \modelName system.

For LR, as shown in Table~\ref{tab:dialogue_metrics1}, our CLIP model with disambiguation outperforms the detector baseline by achieving 100\% success rate in LR and navigation with fewer rounds of dialogue on average. 
We attribute this result to the fact that CLIP is an open vocabulary model that can take free-form query text, which is essential for our task because the user may use different expressions to refer to the same landmark. 
On the contrary, a closed vocabulary object detector can only handle a fixed set of object classes with limited expressions. 
For example, in Table~\ref{tab:grounding}, although both models can handle different objects that belong to the same landmark, CLIP can associate synonyms, such as ``sofa'' and ``couch'', and wrong transcriptions, such as ``coach'', to the correct landmark. 
In contrast, the closed-vocabulary detector can only handle strictly fixed expressions.
The detector misidentifies some objects such as the transparent door in Landmark C after fine-tuning. 
Since our target users are usually non-experts, the baseline sometimes needs the user to rephrase multiple times to recognize the goal, which causes the user to run out of patience, and results in failure or more rounds of dialogue.


Besides CLIP, the disambiguation dialogue also contributes to the performance. With disambiguation, additional information such as the material and functionality of objects can be merged into the query text, such as ``fabric chair'' and ``relaxing chair'' as shown in Table~\ref{tab:grounding}. 
These rich descriptions are helpful in distinguishing landmarks that have the same objects with different attributes, such as the different types of chairs in Landmark A, D, and E in Fig.~\ref{fig:environment} with fewer rounds of user rephrasing. 

\textbf{NLU:}
In Table \ref{tab:dialogue_metrics2}, the overall accuracy of NLU is over 15\% higher than SR, as the NLU is trained with incorrectly transcribed text and thus can work even when SR is incorrect. 
However, we do notice that NLU performs better with correct SR.
The common failure cases of NLU occur when (1) The SR mistakenly breaks a sentence into two halves (\eg\: ``Is there anything?'' and ``To my right.'' are treated as two sentences); and (2) The NLU does not correctly extract intents from noisy transcriptions and chitchat, which can happen during the user study.
Thus, we believe that a better SR engine would vastly benefit the performance of the whole system. 
However, since \modelName will not begin navigation until the user confirms the goal in the dialogue, the wrong SR and NLU have little effect on navigation. 


\textbf{Other Modules:}
The system's environment descriptions are sometimes inaccurate due to errors in the object detector such as: (1) detecting incorrect number of objects (\eg~3 wall sockets, when there was only 1 present); and (2) incorrect object classifications of rare or uncommon objects (\eg~a building information tablet was classified as a poster).
Although we use non-maximum suppression and confidence score threshold to reduce the errors, they are hard to entirely eliminate due to the data distribution shift and the blurry images caused by the robot motion. 
Nevertheless, in Table~\ref{tab:dialogue_metrics2}, the model is able to output a list of objects with correct class names in $75.76\%$ of the cases, which might be more important to the user than a correct number of objects. 

The VQA module accurately answers the user's questions in 82.43\% of the cases.
The model fails in cases where the user asks questions that the robot cannot answer based on a single RGB image. 
For example, without precise depth information the VQA model only answers ``far'' or ``close'' if the question is ``How far is the person from me?''. Without a wider field of view, the model outputs objects on the front side if the question is ``What is on my right?''.

\subsection{Qualitative Evaluation}
In Table \ref{tab:routes}, participants showed an increasing preference for \modelName with CLIP over the detector in all user experience categories across all routes. 
Specifically, participants reported a 32\% improvement with a mean score difference of $1.60 \pm 0.89$ in the overall experience and a mean score difference of $1.40 \pm 0.89$ in the communication experience.
The difference increases as the goal landmark contains more complicated objects in Route 2, and objects that are difficult to detect in Route 3, where the failures in LR significantly lower the user score for the detector based system. 
Particularly, participants noted that \modelName with CLIP understood their intent, asked good follow-up questions, and correctly guided them to their destination. In contrast, the closed-vocabulary detector failed at these aspects and occasionally was unable to recognize destinations even though they existed. 
Participants also noted that the failures in intent understanding led to a frustrating communication experience with the detector. 




One user in particular mentioned that the CLIP based model ``... was able to actually understand me, so it accurately took me to the location and correctly answer~[sic] my questions." while the detector based model ``... would confirm the location I wanted to go to but could not find~[sic; participant meant understand] the right location." 
However, users also mentioned potential improvements for \modelName including more detailed environment descriptions, a quicker response time, and warnings of potential dangers such as ``We're going through a door.''

For the user experience categories that are the same for both LR methods, such as the `intuitiveness of communication interface' and the ability of the system to aid in `gaining awareness of the environment,' participants reported average scores of $7.07 \pm 2.17$ and $6.07 \pm 3.21$, respectively.
As evidenced by these scores, the users' opinions regarding these two categories were positive,  due to the inclusion of the dialogue and grounding modules.
However, participants highlighted minor inaccuracies in the environment descriptions and the slow pace of communication due to processing times and network delays as potential issues.


\begin{table}[t]
    \centering
    \caption{Mean user experience scores on a scale of 1 to 10.}
    \label{tab:routes}
    \resizebox{\linewidth}{!}{\begin{tabular}{l cc cc cc}
        \toprule
         \multirow{2}{*}{\textbf{Use experience category}}& \multicolumn{2}{c}{\textbf{Route 1}} & \multicolumn{2}{c}{\textbf{Route 2}} & \multicolumn{2}{c}{\textbf{Route 3}} \\
         \cmidrule(lr){2-3}\cmidrule(lr){4-5}\cmidrule(lr){6-7}
         & CLIP & Detector & CLIP & Detector & CLIP & Detector \\
         \midrule
         Ease of following & 
         \textbf{8.8} &
         8.6 &
         \textbf{8.8} &
         5.6 &
         \textbf{9.2} &
         1.0
         \\
         Navigational Experience & 
         \textbf{8.4} &
         7.4 &
         \textbf{7.6}  &
         4.8 &
         \textbf{8.8} &
         1.0
         \\
         Intent Understanding & 
         7.6 &
         \textbf{8} &
         \textbf{7.6} &
         4.6 &
         \textbf{8.4} &
         3.4 
         \\
        \bottomrule
    \end{tabular}}
    \vspace{-10pt}
\end{table}

\section{Conclusion and Future Work}
\label{sec:conclusion}
In conclusion, we present \modelName, a first-of-its-kind guide robot that fulfills user intents and familiarizes the user with their surroundings through interactive dialogue. 
We use CLIP to retrieve landmark destinations from commands and provide visual information through language. 
The user study shows promising communication, grounding, and navigation performance of \modelName. 
Our work suggests that visual-language grounding and dialogue can greatly improve human-robot interaction.

To extend \modelName and address its limitations, we point out the following directions for future work. 
First, the current dialogue system is rule-based with fixed behaviors for each intent. 
Replacing hard-coded rules with adaptive learning-based policies, such as large language models, should generalize to more complex user behaviors and more subtasks. 
Second, the environment understanding modules provide limited information. Future informative descriptions should include object relationships in images, incorporate information from the map and other sensors, and inform users about the planned path and potential dangers.
Finally, the physical interface of the platform should be redesigned to improve ergonomics.
\modelName demonstrates the feasibility of vision and language models in assistive navigation that future research in dialogue management, computer vision, and robotics can explore further.

\section*{Acknowledgements} 
The authors would like to thank Samuel Olatunji and Peter Du for their help on the user study, Ye-Ji Mun and Zhe Huang and Neeloy Chakraborty for their feedback on the paper draft, and all participants in the user study.


\bibliographystyle{IEEEtran}
\bibliography{BibFile}

\end{document}